\documentclass[letterpaper]{article}

\usepackage{aaai}
\usepackage{times}
\usepackage{helvet}
\usepackage{courier}

\usepackage{graphicx}
\usepackage{epsfig}
\usepackage{latexsym}
\usepackage{enumerate}
\usepackage{amsmath}
\usepackage{hyperref}
\usepackage{url}

\newcommand{\checkit}[1]{}
 
\newcommand{\Omit}[1]{}

\newcommand{\denselist}{\itemsep -1pt\partopsep 0pt}

\newtheorem{theorem}{Theorem}
\newtheorem{proposition}[theorem]{Proposition}

\newtheorem{definition}[theorem]{Definition}

\newcommand{\calG}{{\cal G}}

\pdfoutput=1
\nocopyright

\title{Heuristics for Planning, Plan Recognition and Parsing \\ (Written: June 2009, Published: May 2016) }
  \author{ Miquel Ram\'{\i}rez  \\
            Departament of Technology (DTIC) \\
             Universitat Pompeu Fabra \\
          08018  Barcelona, SPAIN \\
         {\normalsize \texttt{miquel.ramirez@protonmail.ch}} \\
       \And 
         Hector Geffner \\
          Departament of Technology (DTIC) \\
         ICREA \&  Universitat Pompeu Fabra \\
          08018  Barcelona, SPAIN \\
         {\normalsize  \texttt{hector.geffner@upf.edu}}}
\begin{document}

\maketitle 

\begin{abstract}
In a recent paper, we have shown that Plan Recognition over \textsc{Strips} 
can be formulated and solved using Classical Planning heuristics and algorithms
\cite{ramirez-geffner:ijcai09}. In this work, we show that this formulation subsumes the standard 
formulation of Plan Recognition over libraries  through a compilation of  libraries 
into \textsc{Strips} theories. The libraries correspond to AND/OR graphs 
that may be cyclic and where children of AND nodes may  be partially ordered. 
These libraries include Context-Free Grammars as a special case, where
the Plan Recognition problem becomes a parsing with missing tokens problem.
Plan Recognition over the standard  libraries become
 Planning problems that can be easily solved by 
any modern planner, while recognition over more complex libraries, 
including Context--Free Grammars (\textsc{CFG}s), illustrate limitations of current Planning heuristics
and suggest improvements that may be relevant in other Planning
problems too.
\end{abstract}

\section{Introduction}

Plan Recognition is a common task in a number of areas
where the goal and plan of an agent must
be inferred from observations of its behavior 
\cite{schmidt_sridharan_goodson:aij78,cohen_perrault_allen:81,kautz:aaai06}.
Plan Recognition is a form of Planning in reverse: while in Planning, we seek the actions
that  achieve a goal, in Plan Recognition, we seek the goals that explain the observed actions.
Work in Plan Recognition, however, has proceeded independently of the work in Planning,  
using mostly handcrafted libraries  or  algorithms  not related to
 Planning~\cite{kautz:pr,vilain:aaai90,charniak_goldman:aij93,lesh-etzioni:ijcai95,goldman_geib_miller:uai99,avrahami_kaminka:ijcai05}.%%%

Recently, we have shown  that Plan Recognition  can be formulated
and solved using Classical Planning algorithms \cite{ramirez-geffner:ijcai09}. 
This is important since Classical Planning algorithms have become  quite
powerful in recent years. This formulation does not work over libraries
but over \textsc{Strips} theories where a set ${\cal G}$ of possible goals is given.
The Plan Recognition task  is defined as the problem of identifying the goals $G \in {\cal G}$
that have some \emph{optimal plan} compatible with the observations $O$. Such goals are grouped into the 
\emph{optimal goal set} ${\cal G}^*$, ${\cal G}^* \subseteq {\cal G}$. 
The reason for focusing on the optimal plans is that  they 
represent the possible behaviors of a perfectly rational agent pursuing the goal $G$
\cite{tenenbaum:pr}.  By suitable  transformation, it is then shown in \cite{ramirez-geffner:ijcai09}
that this optimal set ${\cal G}^*$ can be computed \emph{exactly} by means of optimal Planning algorithms 
and \emph{approximately}  by efficient suboptimal Planning algorithms and polynomial
heuristics.

In this work, we show that this formulation subsumes the standard 
formulation of Plan Recognition over libraries through a compilation of  libraries 
into \textsc{Strips}. The libraries correspond AND/OR graphs 
that may be cyclic and where children of AND nodes may  be partially ordered. 
This libraries include Context-Free Grammars as a special case, where
the Plan Recognition problem becomes a  parsing problem.
Plan Recognition over the standard Plan Libraries become
simple Planning problems that can be easily solved by 
any modern planner, while recognition over more complex libraries, 
including \textsc{CFG}s, illustrate limitations of current Planning heuristics
and improvements that may be relevant in other Planning problems as well.\footnote{
Parsing in \textsc{CFG}s is polynomial  while Planning is known to be \textsc{NP}--hard. This worst complexity
bounds, however, do not imply that the reduction of parsing to Planning is necessarily
a bad idea. First, many Planning problems -- like many\textsc{SAT} problems --
can be solved quite efficiently; second, parsing with constraints, 
as required in Natural Language Processing, is also intractable,
yet many of these constraints can be handled naturally in Planning.
In addition, the mapping handles
missing tokens in the input sentence and yields interesting
lessons for Planning heuristics.}

The paper is organized as follows. First we review the formulation of Plan Recognition
over \textsc{Strips} theories in \cite{ramirez-geffner:ijcai09}, then we consider Plan Recognition
over libraries, present some experimental results, and draw some conclusions.

\section{Plan Recognition as Planning} 

A \textsc{Strips}  Planning problem  is a tuple $P=  \langle F,I,A,G \rangle$
where $F$ is   the set of   fluents, $I \subseteq F$ and $G \subseteq F$ are the initial and goal situations,
and $A$ is a set of actions $a$ with precondition, add, and delete lists $Pre(a)$, $Add(a)$, and $Del(a)$ respectively, all of
which are subsets of $F$. For each action $a \in A$, we assume  that there is a \emph{non-negative cost}  $c(a)$  so that the cost 
of a sequential plan $\pi = a_1, \ldots, a_n$ is  $c(\pi) = \sum  c(a_i)$. A plan $\pi$ is \emph{optimal}
if it has minimum cost.  For unit costs, i.e., $c(a)=1$ for all $a \in A$, plan cost is plan length, and
the optimal plans are the shortest ones. Unless stated otherwise, action costs are assumed to be $1$.

\subsection{Definition}

The Plan Recognition problem given a \emph{plan library} $L$ for a set $\cal G$ of possible goals $G$
can be understood, at an abstract level, as the problem of finding a goal $G$ with a plan $\pi$
in the library, written $\pi \in \Pi_L(G)$, such that $\pi$  satisfies the observations. 
We define the Plan Recognition problem over a \emph{domain theory} in a similar way just
changing the set $\Pi_L(G)$ of plans for $G$ in the library by the set $\Pi^*_P(G)$ of optimal
plans for $G$ given the  domain $P$.  We use $P =\langle F,I,O\rangle$  to represent  Planning \emph{domains}
so that a Planning \emph{problem} $P(G)$ is obtained by concatenating a Planning domain with a goal $G$, 
which is a set of fluents. We define a \emph{Plan Recognition problem} or \emph{theory} as follows:

\begin{definition}
A Plan Recognition problem or \emph{theory} is a triplet  $T = \langle P, \calG, O \rangle$ where $P=  \langle F,I,A\rangle$ is a Planning domain, 
$\cal G$ is the  set of possible goals $G$, $G \subseteq F$, and $O = o_1, \ldots, o_m$ is an \emph{observation sequence} with each $o_i$ being an action in $A$.
\end{definition}

We also need to make precise what it means for an action sequence to satisfy an observation sequence made up of actions.
E.g., the action sequence  $\pi=\{a,b,c,d,e,a\}$ satisfies the observation sequences 
$O_1 = \{b,d,a\}$ and $O_2=\{a,c,a\}$,  but not  $O_3  = \{b,d,c\}$. This can be formalized 
with the help of a function that maps observation indices in $O$ into action indices in $A$:
\begin{definition} 
An action sequence $\pi = a_1, \ldots, a_n$ satisfies the  observation sequence  $O = o_1, \ldots, o_m$ if there is a \emph{monotonic} function $f$
mapping the observation indices $j=1, \ldots, m$ into action indices $i=1, \ldots, n$, such that $a_{f(j)}=o_j$.
\end{definition}
%
% For example,  the unique function $f$ that establishes a correspondence between the actions  $o_i$ observed  in $O_1$ 
% and the actions $a_j$ in $\pi$ is   $f(1)=2$, $f(2)=4$, and $f(3)=6$. This function must be strictly 
% monotonic so that the action sequences $\pi$ preserve the ordering of the actions observed.

The solution to a Plan Recognition theory $T=\langle P,\calG, O\rangle$ is given by the goals $G$ 
that admit an optimal plan that is compatible with the observations:
\begin{definition}
The exact solution to a  theory  $T=\langle P, \calG, O\rangle$ is given by the 
\emph{optimal goal set} ${\cal G}_T^*$ which comprises the  goals $G \in {\cal G}$
such that for some $\pi \in \Pi_P^*(G)$, $\pi$ satisfies $O$.
\end{definition}
%

%We say also  that those goals $G$ are the ones that  explain  or account  for  the observations.

%
% \begin{figure}[htbp]
\begin{figure}
\centering
\includegraphics[width=0.4\linewidth,keepaspectratio]{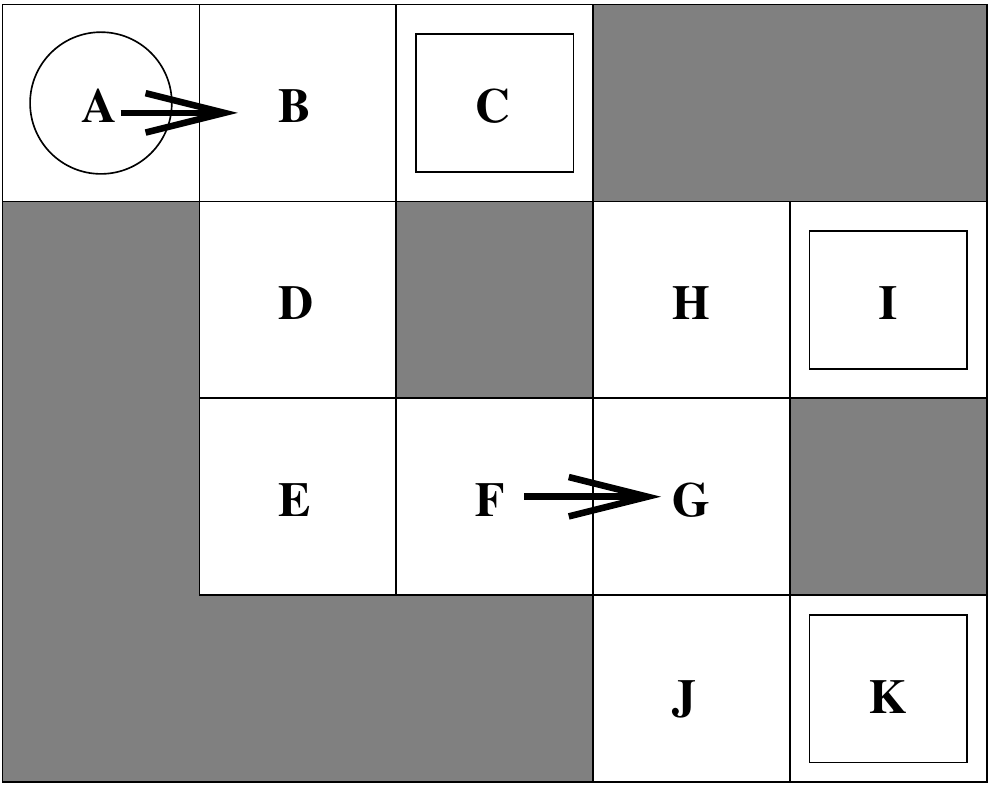}
\caption{\small Plan Recognition: Is the agent headed to $C$, $I$, or $K$? The observations are the  transitions
from $A$ to $B$ and $F$ to $G$ in that order.}
\label{fig:grid}
\end{figure}

Figure~\ref{fig:grid} shows a simple Plan Recognition problem.  Room A (marked with a circle) is the initial position of the agent, while 
Rooms C, I and K (marked with a  square) are its  possible  destinations. Arrows between Rooms A and B, and F and G, are the observed agent movements
in that order. In the resulting theory $T$, the only possible goals that have optimal plans compatible with the observation sequence 
are I and K. In the terminology above, the set of possible goals  ${\cal G}$ is given by the  atoms $at(C)$, $at(I)$, and $at(K)$, 
while the optimal goal set ${\cal G}_T^*$  comprises  $at(I)$ and $at(K)$,  leaving out the possible 
goal $at(C)$.

\subsection{Computation}

In order to solve the Plan Recognition problem  using Planning algorithms, we get rid of   the observations.
For simplicity, we  assume that no pair of observations $o_i$ and $o_j$ refer to the same action $a$ in $P$. When this is not so, 
we create a copy $a'$ of the action  $a$ in $P$ so that $o_i$ refers to $a'$ and $o_j$ refers to $a$.

We will eliminate observations  by mapping the theory $T=\langle P,\calG, O\rangle$ into an slightly
different theory $T'=\langle P',{\cal G}', O'\rangle$ with an \emph{empty} set $O'$ of observations, such that  the
solution set ${\cal G}^*_T$ for $T$ can be read off from the solution set ${\cal G}^*_{T'}$ for $T'$.
\begin{definition}
For a theory $T=\langle P,{\cal G}, O\rangle$, the transformed theory is $T'=\langle P',{\cal G}', O'\rangle$ with
\begin{itemize} \denselist
\item $P'= \langle F', I', A'\rangle$ has fluents $F' = F \cup F_o$, initial situation $I' = I$, 
and actions $A' =  A \cup A_o$,  where  $P= \langle F, I, A\rangle$, 
$F_0 = \{p_a \ | \ a \in O\}$, and $A_o = \{o_a \ | \ a \in O\}$, 
\item ${\cal G}'$ contains the  goal $G' = G \cup G_o$ for each goal $G$ in $\cal G$, where 
$G_o = F_o$, 
\item $O'$ is empty
\end{itemize}
\label{def:transf}
\end{definition}
The new actions  $o_a$ in $P'$ have   the same precondition, add, and delete lists as the actions $a$ in $P$ except for 
the new fluent $p_a$ that is added to $Add(o_a)$, and the fluent  $p_b$,  for the  action  $b$ that immediately  precedes  $a$ in $O$,  if any, 
that is  added to $Pre(o_a)$.

In the transformed theory $T'$, the observations $a \in O$ are  encoded as extra fluents $p_a \in F_o$, extra
actions $o_a \in A_o$, and extra goals $p_a \in G_o$. Moreover, these extra goals $p_a$ can only be achieved by the new actions $o_a$, 
that due to the precondition $p_b$ for  the  action $b$ that precedes $a$ in $O$,   can be applied only after all 
the actions  preceding $a$ in $O$,  have been executed.  The result is that the  plans 
that achieve the  goal $G' = G \cup G_o$  in $P'$ are in correspondence with the plans that achieve the goal $G$
in $P$ \emph{that satisfy the observations} $O$:

\begin{proposition}
$\pi = a_1, \ldots, a_n$ is a plan for $G$ in $P$ that satisfies the observations $O=o_1, \ldots,o_m$ under the 
function $f$ iff $\pi'=b_1, \ldots, b_n$ is a plan for $G'$ in $P'$ with  $b_i=o_{a_i}$,  if $i=f(j)$ for some $j \in [1,m]$, 
and $b_i=a_i$ otherwise.
\label{obs}
\end{proposition}

It follows from this that $\pi$ is an optimal plan for $G$ in $P$ that satisfies the observations
iff $\pi'$ is an optimal plan in $P'$ for two different goals: $G$, on the one hand, and $G'=G \cup G_o$
on the other. If we let $\Pi_P^*(G)$ stand for the set of optimal plans for $G$ in $P$, we can 
thus  test whether a goal $G$ in $\cal G$ accounts for the observation as follows:\footnote{
Note that while a plan for $G' = G \cup G_o$ is always a plan for $G$, it is \emph{not} true
that an \emph{optimal} plan for $G'$ is an \emph{optimal} plan for $G$, or even  that a 
good plan  for $G'$ is a  good plan  for $G$.}
\begin{theorem}
$G \in {\cal G}_T^*$ iff there is an   action sequence  $\pi$ in $\Pi_{P'}^*(G) \cap \Pi^*_{P'}(G')$.
\label{thm:transf}
\end{theorem}
Moreover, since $G \subseteq G'$, if we let $c^*_{P'}(G)$ stand for the optimal cost of achieving $G$ in $P'$, 
we can state this result in a simpler form:
\begin{theorem}
$G \in {\cal G}_T^*$ \ iff \ $c^*_{P'}(G) = c^*_{P'}(G')$
\label{thm:2}
\end{theorem}

% We will use Theorem~\ref{thm:2} below for computing the optimal goal  set ${\cal G}^*_T$ exactly using an slightly modified optimal planner, 
% and then make use of Theorem~\ref{thm:transf}, for motivating methods that approximate ${\cal G}_T^*$ and scale up better. 
%

% As an example of the transformation, for the plan recognition task shown  in Figure~\ref{fig:grid}, the extra fluents $F_o$ in $T'$
% are $p_{move(A,B)}$ and $p_{move(F,G)}$, while the extra actions $A_o$ are $o_{move(A,B)}$ and $o_{move(F,G)}$; the first
% with the same precondition as $move(A,B)$ but with the extra fluent $p_{move(A,B)}$ in the Add list, 
% and the second    with the extra  precondition $p_{move(A,B)}$ and the extra effect $p_{move(F,G)}$.
% Theorem~\ref{thm:2} then means that a goal $G$ accounts for the observations in the original theory $T$,
% and thus belongs to ${\cal G}_T^*$, iff the cost of achieving the goal $G$ in the transformed domain $P'$ 
% is equal to the cost of achieving the goal $G' = G \cup G_o$. \emph{In the transformed problem, thus, 
% observations have been replaced by extra goals that must be achieved at no extra cost.}

The optimal goal set ${\cal G}^*$ can be computed, using this result, by solving two optimal Planning problems for 
each possible goal $G$: one extending the domain $P'$ with the goal  $G$, the other extending $P'$ with the goal $G'$
made up of $G$ and the dummy goals $G_o$ encoding the observations. The goal $G$ explains the observations
and thus belongs to ${\cal G}^*_T$ iff the solutions to these two optimal Planning problems have the same cost.
In \cite{ramirez-geffner:ijcai09}, a more efficient method for computing this set exactly  is introduced, where the cost of the first problem is used as the upper bound in the solution of the second. In addition, two methods
that approximate ${\cal G}_T^*$  and scale up much better are presented. For Plan Recognition
over libraries, the situation is simpler, as the resulting Planning problems have \emph{zero action costs},
and hence {\em all plans are optimal.}

\section{Plan Recognition over Libraries}

As mentioned above, the Plan Recognition problem given a \emph{plan library} $L$ for a set $\cal G$ of possible goals $G$
can be understood, at an abstract level, as the problem of finding a goal $G$ with a plan $\pi \in L$, 
written $\pi \in \Pi_L(G)$, such that $\pi$  satisfies the observations $O$. We show now that a library $L$ for a goal $G$
can be compiled intro a \textsc{Strips} Planning problem $P_L(G)$ so that $\pi$ is in $\Pi_L(G)$ iff $\pi$ is a plan for $P_L(G)$.
Provided with this correspondence and {\em by setting the cost of all the actions in $P_L$ to zero}, so that
no plan in the library is ruled out due to their cost, a plan in the library for $G$ will satisfy
the observations $O$ iff $G$ is in the optimal goal set ${\cal G}_{T}^*$ of the theory $T = \langle P_L, {\cal G}, O\rangle$,
a set that can be computed by using an off-the-shelf classical planner upon
the Planning problems $P'_L(G')$ obtained from the transformation that compiles the observation $O$ in $T$
away.

\subsection{Plan Libraries}

As it is standard, we take a library $L$ for a goal $G$ to be a rooted, ordered  AND/OR graph 
where each node is a AND node, an OR node, or  a leaf.  Leaves represent primitive task (actions), 
OR nodes represent non-primitive tasks,  and AND nodes represent methods for decomposing non-primitive
task. The children of OR nodes are AND nodes or leaves, while the children of AND nodes 
are OR nodes or leaves.  The children of an  AND node $n$  can be  ordered partially;
we write $n' <_n n''$ to express that child $n'$ of $n$ must come before child $n''$.
The root of the library is a task (OR node) that represents the goal $G$ to be achieved. 
We will allow libraries to be cyclic, and thus, \textsc{CFG}s will 
be an special case  where the OR nodes stand for the non-terminal symbols in the grammar, the AND nodes stands for the grammar
rules, and the leaves stand for the grammar terminals. The children of the AND/OR graphs
that represent \textsc{CFG}s are normally cyclic and the children of AND nodes (rules) are 
ordered linearly.

The set of solutions to one such AND/OR graph can be defined by means of derivations
as it is common in parsing, with the only difference that a partially ordered
rule $X \rightarrow Y_1, \ldots, Y_m$ represented by an AND node, stands for
the set of all totally ordered rules $X \rightarrow Y_{i_1}, \ldots, Y_{i_m}$
compatible with the partial order. The set of plans $\Pi_L(G)$ in the library
for $G$ denotes the set of 'strings' (sequences of terminal tasks or actions) 
that can be derived from the root node corresponding to $G$.

\subsection{Compilation}

The compilation of the library $L$ for a goal $G$ into a \textsc{Strips} Planning problem $P_L(G)$ 
depends on a depth parameter $N$, and it ensures that the plans in $P^N_L(G)$ are in correspondence with the set of plans (strings of primitive tasks) $\Pi_L(G)$ that can be derived from the 
library by bounding the depth of the derivation to $N$. If the library is acyclic,
it suffices to set $N$ to the depth of the graph to ensure completeness; otherwise, 
the parameter $N$ puts a bound on the number of derivations. For simplicity,
we often drop the index $N$ from the notation.

The Planning problem $P_L(G) = \langle F_L, I_L, G_L, A_L\rangle$
have a set of fluents $F_L$, initial and goal situations $I_L$ and $G_L$, 
and actions $A_L$. For simplicity, we will describe the problem assuming
 a \textsc{Strips} language with negation. Negation,
however, can be easily compiled away \cite{gazen:adl}. 

The fluents $F_L$ in $P_L$ are $started(n,i)$, $\neg started(n,i)$, $finished(n,i)$, $\neg finished(n,i)$,
and $top(i)$, where $n$ corresponds to the nodes in the AND/OR graph representing
the library $L$, and $i=[0 \ldots  N]$. The integers $i$ aim to capture the possible levels of the stack,
with the true level captured by the fluent $top(i)$ that is mutex with $top(k)$ for $k\not= i$. 
In a state, where $top(i)$ is true, the fluents $started(n,k)$ and $finished(n,k)$ for $k \leq  i$
express the contents of the stack. In any such a state, all fluents $started(n,k)$ and $finished(n,k)$ for $k > i$
will be false.

The initial and goal situations of $P_L$ are  $I_L = \{top(0)\}$ and 
$G_L = \{finished(n,0)\}$, where $n$ is the single (OR) root node of the library $L$.
That is, the stack starts at level $0$ empty with no node started, 
and the goal is to finish with the root node executed at the same level.
For doing this, the stack will expand and contract, while 
the execution of a node will allow the execution of its children.
Roughly the $started(n,i)$ fluents flow downward in the graph, 
and  the fluents $finished(n,i)$ flow upward, with the actions
$start(n,i)$ and $end(n,i)$ emulating the start and ending
of the primitive and non-primitive tasks in the AND/OR graph.
As a convenient abbreviation, we write $i \! + \! 1$ and $i \! - \! 1$ to denote constants $i'$
defined as the successor and predecessor of the constant $i$ in the
encoding. The actions in $P_{L}^{N}(G)$are:

{\small
\begin{itemize}
\item Calls from \textsc{And} nodes $n$ to non--terminal children $n'$ are represented by actions $start(n,n',i)$ with preconditions
\[
\begin{split}
Pre &= \{ top(i), started(n,i), \neg finished(n') \} \\
  & \cup \{ finished(n'', i) \, | \, n'' <_{n} n'\} \\
 & \cup \{ \neg started(n'', i) \, | \, n'' \in children(n) \}
\end{split}
\]
add list $Add = \{ top(i\! + \!1), started(n',i\! + \!1\}$ and delete list $Del = \{top(i)\}$. For calls to terminal children $n'$, the precondition of $start(n,n',i)$ is the same as the one described above but $Add = \{ finished(n',i) \}$ and $Del = \emptyset$. 
\item Termination of calls made from \textsc{And} nodes $n$ are encoded with actions $end(n,i)$, with preconditions
\[
\begin{split}
Pre &= \{ top(i), started(n,i) \} \\
    & \cup \{ finished(n') \, | \, n' \in children(n) \}
\end{split}
\]
and add list $Add = \{ finished(n,i-1), top(i-1) \}$ and delete list $Del = Pre$.
\item Calls from internal \textsc{Or} nodes $n$ to children $n'$ are represented by actions $start(n,n',i)$ with precondition
\[
\begin{split}
Pre &= \{ top(i), started(n,i) \} \\
    &\cup \{ \neg finished(n'', i\! + \!1) \, | \, n'' \in children(n) \} \\
    &\cup \{ \neg started(n'', i\! + \!1) \, | \,  n'' \in children(n) \} 
\end{split}
\]
add list $Add = \{ top(i\! + \!1), started(n', i\! + \! 1) \}$ and delete list $Del = \{ top(i) \}$. 
\item Termination of calls from internal \textsc{Or} nodes are represented by actions $end(n,n',i)$, where $n'$
is a child of $n$, with precondition $Pre = \{ top(i), started(n,i), finished(n',i) \}$, add list 
$Add = \{ finished(n, i\! - \!1), top(i\! - \!1) \}$ and delete list $Del = Pre$.
\item 
Root \textsc{Or} nodes are handled like other OR nodes, except that the action $end(n,i=0)$ 
adds $finished(n,0)$ rather than adding $finished(n,i\! - \!1)$.

\end{itemize}
}

For a plan $\pi$ for $P_L(G)$, let us keep only the sequence of $start(n,n',k)$ actions where $n$ is a leaf node of $L$,
and let us set $f_L(\pi)$  to the corresponding sequence with the $start(n,n',k)$ actions replaced
by the primitive actions associated with the nodes $n$. 

The first result is about the  correspondence between the set of plans in the  library $L$ for $G$ with
depth bounded by $N$, $\Pi_L^N(G)$, and the sequences of primitive actions $f(\pi)$ for plans $\pi$ for
$P_{L}^{N}(G)$:

\begin{theorem}[Correspondence]
For a library $L$ for goal $G$   and a positive integer  $N$, 
 $\pi \in \Pi_L^N(G)$ iff there is a  plan $\pi'$ for $P_L(G)$ such that $\pi =  f_L(\pi)$.
\end{theorem} 

The second result exploits this correspondence for computing the plans
in the library that comply with a set of observations $O$ using an off-the-shelf
classical planner, suboptimal or not, over the problem $P'_L(G')$ obtained
from $P_L(G)$ by compiling the observations $O$ away (Definition~\ref{def:transf}):

\begin{theorem}[Computation]
For a library $L$ for a goal $G$,  and a positive integer $N$, 
$G$ has a plan  in $\Pi_L^N(G)$ compatible with the observations $O$ iff there is a plan
for the Planning problem ${P'}_L^N(G')$ obtained from $P_L(G)$ by compiling
the observations $O$ away.
\end{theorem}

The third result is semantic, and shows that this computational method
follows from the general formulation for Plan Recognition from
\textsc{Strips} theories when action costs are taken to be zero:

\begin{theorem}[Subsumption]
Let ${\cal G}$ be a set of possible goals, and let $L$ be the library
for $G \in {\cal G}$. Then $G$ has a plan in the library that satisfies
the observations $O$ with depth no greater than $N$ iff there is
an \emph{optimal plan} for the problem $P^N_L(G)$ that satisfies $O$,
assuming action costs to be zero.
\end{theorem}

Indeed, this result follows from the one above, as when all action
costs are zero, \emph{any plan} for ${P'}^N_L(G')$ is an \emph{optimal plan}
for ${P'}^N_L(G)$, which in turn from Proposition~\ref{obs}, represents
a plan for the problem $P_L(G)$ that satisfies the observations.
% 
% 
% 
% ** Question: after compilation, what happens if we add to possible goal set ${\cal G}$,
% \emph{sets of goals}, e.g. $G_1$ and  $G_2$, each with its own AND/OR graph. Composed problem
% $P_L(G_1) \cup P_L(G_2)$ seems to define a Strips problem -- join I's and G's,  etc.
% Does resulting Strips theory handle 'interleaving' as Geib wants?
% What if there are primitive actions in common? It looks that correspond
% should hold too (or not?). Check ***
\section{Experimental Results}
We test below the Plan Recognition framework laid out above over plan libraries and Context-Free Grammars.

\subsection{The Soccer Plan Library}
\begin{figure*}[htbp]
\centering
\includegraphics[width=0.6\linewidth,keepaspectratio]{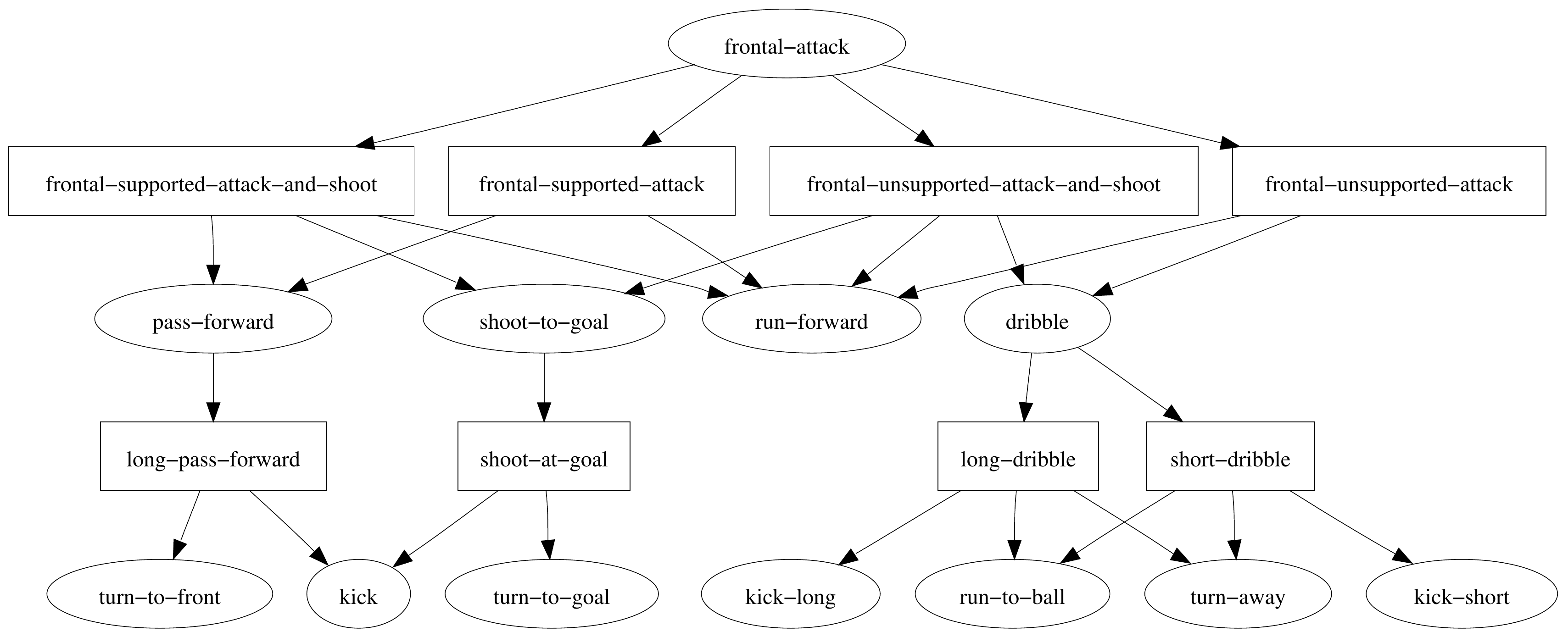}
\caption{{\small Plan library for \emph{Frontal--Attack}. Nodes with elliptic shape are OR nodes or leaves (primitive tasks), box--shaped nodes are AND nodes. Precedence constraints between children of AND nodes are not shown.}} 
\label{fig:frontal-attack}
\end{figure*}

From the descriptions found~\cite{tambe:isis-teamwork} on plan hierarchies for controlling simulated RoboSoccer teams, we have defined ourselves a set plan libraries for recognizing the intentions of the opposing soccer team. Each library considers one of the following four root tasks, namely, \emph{Frontal--Attack}, \emph{Flank--Attack}, \emph{Fight--Back} and \emph{Fall--Back}~\footnote{Names of tasks loosely correspond with those of top--level goals featured by the plan hierarchy found in the ISIS source distribution.}. Plans in the first two libraries share a substantial amount of activities, e.g. kicking the ball, or running towards the general direction of the opposing team, which do not or hardly take place in plans conveyed by the latter two libraries. In Figure~\ref{fig:frontal-attack} we show the plan library for the task \emph{Frontal--Attack}.

In the experiments, we test which plan libraries are compatible with a sequence of observations drawn
from a plan obtained from one of them. The planner we used to search for such plans is the satisfying classical planner \textsc{FFv2.3}. 
% When an observed activity is not covered by library $L$~\footnote{We borrow the term ``cover'' from the \textsc{NLP} jargon, where a grammar $G$ \emph{covers} a terminal $x$ if $x$ appears in some right--hand side of a production of $G$.} the hypothetic goal corresponding to having executed the root task of $L$ is ruled out. 
\begin{table}
\begin{center}
{\tiny
\begin{tabular}{|c|c|c|c|c|c|}
\hline 
Sequence & $|O|$ &  Time & Expanded & Plan Len & Algorithm \\
\hline
\#1 & 2 & 0.2 & 32 & 12 & \textsc{Ehc, Bfs} \\
\#2 & 3 & 0.08 & 139 & -- & \textsc{Bfs} \\
\#3 & 6 & 0.08 & 172 &  -- & \textsc{Bfs} \\
\#4 & 5 & 0.13 & 112 &  -- & \textsc{Bfs} \\
\hline
\end{tabular}
}
\end{center}
\caption{{\small Average time, number of nodes expanded for determining that a library was compatible with the observations, length of resulting explanations (plans) and algorithm used by \textsc{FF} -- either \textsc{Ehc} or \textsc{Bfs}. $|O|$ is the size of the observation sequence. Derivation depth $N$ was set to $5$. Sequence \#1 is $\{$run-forward, kick$\}$, sequence \#2 is $\{$run-forward, turn-to-player, kick$\}$, sequence \#3 is $\{$run-forward, turn-away, kick-short, run-to-ball,turn-to-goal, kick $\}$ and sequence \#4 consists in repeating five times the activity kick. None of the plan libraries account for the observation sequences \#2, \#3 and \#4.}}
\label{table:soccer-pr-data}
\end{table}

In Table~\ref{table:soccer-pr-data} we can see that the Planning problems we obtain from our compilation are handled easily by \textsc{FF}. It is important to note that the size of the observation sequence $|O|$ does not seem to be related with the running time. While the plan library depicted in Figure~\ref{fig:frontal-attack} might be very simple, it is not simpler than the plan libraries typically found in the Plan Recognition literature. 

\subsection{Context-Free Grammars}
Context-free grammars (\textsc{CFG}s) appear to present more interesting Planning challenges than the common plan libraries. First, most, if not all, \textsc{CFG}s of interest in Natural Language Processing (NLP) are cyclic, though in languages like English, the depth of the derivation is not big. On the other hand, \textsc{CFG}s used as benchmarks for parsers, like \textsc{ATIS--3} or \textsc{CommandTalk}~\footnote{These grammars can be found in the \textsc{NLTK} (\url{http://www.nltk.org}) Natural Language Processing toolkit corpora.}, feature thousands of rules. Compiling such grammars into Plan Libraries results in graphs with several thousand nodes. Until recently~\cite{geib_goldman:aij09}, there has not been any serious attempt at developing a set of challenging benchmarks for Plan Recognition algorithms.

We have tested our compilation in a toy \textsc{CFG} of the English language, described below:
{\small
\begin{enumerate} \denselist
\item $S \rightarrow NP\,\,VP$
\item $VP \rightarrow V\,\,NP | V | VP\,\,PP$
\item $NP \rightarrow Det\,\,N | Name | NP\,\,PP$
\item $PP \rightarrow P\,\,NP$
\item $V \rightarrow saw|ate|ran$
\item $N \rightarrow boy|cookie|table|telescope|hill$
\item $Name \rightarrow Jack|Bob$
\item $P \rightarrow with|under$
\item $Det \rightarrow the|a|my$
\end{enumerate}
}
Compiling this simple \textsc{CFG} yields a Plan Library with 85 nodes, which in turn yields a Planning problem with about $800$ actions after having fixed the maximum derivation depth to $10$.

\begin{table}
\begin{center}
{\tiny
\begin{tabular}{|c|c|c|c|c|}
\hline 
Sentence & Time (secs) & Expanded & Plan Len & Algorithm \\
\hline
``Jack ate my cookie'' & 0.31 & 211 & 35 & \textsc{Bfs} \\
``ran the boy under the hill'' & 0.36  & 452 & 67 & \textsc{Bfs} \\
``Jack my cookie'' & 0.32 & 168 & 35 & \textsc{Bfs}\\
``the boy under the hill with my cookie ran'' & 0.44 & 1,722 & -- & \textsc{Bfs}\\
\hline
\end{tabular}
}
\end{center}
\caption{{\small Time needed by \textsc{FF} to accept or reject input sentences, number of nodes expanded in each case, length of explanations (plans) and algorithm used by \textsc{FF} to find the solution. Note that sentences \#2 and \#3 are incomplete. In sentence \#2 the sentence subject is missing, and in sentence \#3 the verb is missing. There is no plan accounting for sentence \#4.}}
\label{table:grammar-pr-data}
\end{table}

Table~\ref{table:grammar-pr-data} confirms our intuition that even very simple \textsc{CFG}s yield significantly more challenging Planning problems than Plan Libraries do. In general more search is required to find a parse tree for the input token sequence. One very interesting property inherent to our approach is its ability to ``interpolate'' missing tokens from the input sentence, as is the case of sentence \#3. In that sentence there is no verb, and the planner introduces one of the available productions for non--terminal $V$ in order to obtain a correct parse tree. In sentence \#2 the subject is missing, and in this case the planner introduces a noun--phrase.

Encouraged by these results, we wanted to conduct an experiment with a ``real grammar''. We aimed at obtaining a parse for sentences using the \textsc{ATIS--3} benchmark \textsc{CFG}. Yet this grammar contains over 3,000 different production rules, which resulted in an AND/OR graph with over 6,000 nodes. The Planning problem resulting from compiling that graph featured over 300,000 actions and a disk footprint of about 2 Gigabytes.

We have thus tested our Plan Recognition framework over a \textsc{CFG} not as complex as \textsc{ATIS--3} but a bit  more complex than the toy \textsc{CFG} above. This second grammar features a much richer lexicon: 7 verbs with tenses and number, over twenty nouns, pronouns, auxiliary verbs and all of English prepositions. It also features rules for modeling pragmatics -- statements, questions and commands -- and taking as well into account applicable syntactic cases -- declarative, imperative and interrogation -- for each pragmatic. This second grammar, when compiled, resulted in an \textsc{AND/OR} graph with 251 nodes, which, after fixing the derivation depth $N$ to 30 to ensure solubility, resulted in a Planning problem with over 10,000 actions.

\begin{table}
\begin{center}
{\tiny
\begin{tabular}{|c|c|c|c|c|}
\hline 
Sentence Type & Time (secs) & Expanded & Plan Len & Algorithm \\
\hline
Covered & 266.5, TO(1) & 1,698 & 54.2 & \textsc{BFS} \\
Incomplete & 271 & 393 & 34 & \textsc{BFS} \\
Not Covered & TO & 4,000 & -- & -- \\
\hline
\end{tabular}
}
\end{center}
\caption{{\small Average time, number of nodes expanded -- for timeouts an educated guess is provided -- and plan lengths obtained with the second grammar. TO stands for timeout (time limit was set to 600 seconds). Twelve sentences were divided into three sets. The \emph{Covered} set contained full sentences covered by the grammar. The \emph{Incomplete} set contained covered sentences with missing tokens. The final set, \emph{Not Covered}, refers to non--English sequences of tokens. }}
\label{table:grammar-pr-data-2}
\end{table}

The results of applying our scheme to this second grammar are shown in Table~\ref{table:grammar-pr-data-2}, where three types of sentences are considered. Interestingly \textsc{FF}, solved pretty well the sentences in the \emph{Covered} set, but had trouble processing the non--English token sequences in the \emph{Not Covered} set. The timeout we get in the \emph{Covered} set corresponds to the sentence ``why did you take the book'', while the sentence ``take the book'' was solved after having to expand just 441 nodes. This observation and the fact that incomplete sentences are much smaller than the average sentence in the \emph{Complete} set, leads us to conclude that in the context of parsing as Planning, the length of the sentence to parse seems to be relevant for the hardness of the problem. We can also see that the ``interpolating'' behavior of our scheme is biased towards providing a reasonably sized parse tree. It is also worthy to note that none of the problems was solved with the incomplete \textsc{EHC} procedure.

The result confirms that the search for plans in the resulting theories becomes much more expensive due to the limitations of current heuristics that make planners like FF get lost in much larger search spaces. Moreover, we have found that it is possible to incorporate some ideas from parsing algorithms like \textsc{CYK}~\cite{cyk} into relaxed--plan graph heuristics, while keeping the heuristic itself computable in polynomial time. We think that such heuristics will help the search to become more focused. Interestingly, the new heuristic is general and thus applies to Planning problems that are completely unrelated to parsing. Unfortunately, we haven't had the time to test these ideas yet, but would like to do that for the camera--ready version if the paper is accepted for the workshop.
\section{Discussion}
We have shown that the framework for plan
recognition over \textsc{Strips} theories, formulated recently in \cite{ramirez-geffner:ijcai09}, 
subsumes the Plan Recognition problem over libraries, as they can be compiled
into \textsc{Strips}. We have also shown that recognition over 
standard libraries become Planning problems that can be easily solved by 
modern planners, while recognition over more complex libraries, 
including \textsc{CFG}s, illustrate limitations of current Planning heuristics
and suggest improvements that may be relevant in other Planning
problems as well (to be worked out and shown).

\bibliographystyle{aaai}
\bibliography{control}

\begin{thebibliography}{}

\bibitem[\protect\citeauthoryear{Avrahami-Zilberbrand and
  Kaminka}{2005}]{avrahami_kaminka:ijcai05}
Avrahami-Zilberbrand, D., and Kaminka, G.~A.
\newblock 2005.
\newblock Fast and complete symbolic plan recognition.
\newblock In {\em Proceedings of IJCAI},  653--658.

\bibitem[\protect\citeauthoryear{Baker, Tenenbaum, and
  Saxe}{2007}]{tenenbaum:pr}
Baker, C.~L.; Tenenbaum, J.~B.; and Saxe, R.~R.
\newblock 2007.
\newblock Goal inference as inverse planning.
\newblock In {\em Proceedings of the Twenty-Ninth Annual Conference of the
  Cognitive Science Society}.

\bibitem[\protect\citeauthoryear{Charniak and
  Goldman}{1993}]{charniak_goldman:aij93}
Charniak, E., and Goldman, R.~P.
\newblock 1993.
\newblock A bayesian model of plan recognition.
\newblock {\em Artificial Intelligence} 64:53--79.

\bibitem[\protect\citeauthoryear{Cohen, Perrault, and
  Allen}{1981}]{cohen_perrault_allen:81}
Cohen, P.~R.; Perrault, C.~R.; and Allen, J.~F.
\newblock 1981.
\newblock Beyond question answering.
\newblock In Lehnert, W., and Ringle, M., eds., {\em Strategies for Natural
  Language Processing}. Lawrence Erlbaum Associates.

\bibitem[\protect\citeauthoryear{Gazen and Knoblock}{1997}]{gazen:adl}
Gazen, B., and Knoblock, C.
\newblock 1997.
\newblock Combining the expressiveness of {U}{C}{P}{O}{P} with the efficiency
  of {G}raphplan.
\newblock In Steel, S., and Alami, R., eds., {\em Recent Advances in AI
  Planning. Proc. 4th European Conf. on Planning (ECP-97). Lect. Notes in AI
  1348},  221--233.
\newblock Springer.

\bibitem[\protect\citeauthoryear{Geib and Goldman}{2009}]{geib_goldman:aij09}
Geib, C.~W., and Goldman, R.~P.
\newblock 2009.
\newblock A probabilistic plan recognition algorithm based on plan tree
  grammars.
\newblock {\em Artificial Intelligence} 173:1101--1132.

\bibitem[\protect\citeauthoryear{Goldman, Geib, and
  Miller}{1999}]{goldman_geib_miller:uai99}
Goldman, R.~P.; Geib, C.~W.; and Miller, C.~A.
\newblock 1999.
\newblock A new model of plan recognition.
\newblock In {\em Proceedings of the 1999 Conference on Uncertainty in
  Artificial Intelligence}.

\bibitem[\protect\citeauthoryear{Kautz and Allen}{1986}]{kautz:pr}
Kautz, H., and Allen, J.~F.
\newblock 1986.
\newblock Generalized plan recognition.
\newblock In {\em AAAI},  32--37.

\bibitem[\protect\citeauthoryear{Lesh and Etzioni}{1995}]{lesh-etzioni:ijcai95}
Lesh, N., and Etzioni, O.
\newblock 1995.
\newblock A sound and fast goal recognizer.
\newblock In {\em Proc. IJCAI-95},  1704--1710.

\bibitem[\protect\citeauthoryear{Pentney \bgroup et al.\egroup
  }{2006}]{kautz:aaai06}
Pentney, W.; Popescu, A.; Wang, S.; Kautz, H.; and Philipose, M.
\newblock 2006.
\newblock Sensor-based understanding of daily life via large-scale use of
  common sense.
\newblock In {\em Proceedings of AAAI}.

\bibitem[\protect\citeauthoryear{Ramirez and
  Geffner}{2009}]{ramirez-geffner:ijcai09}
Ramirez, M., and Geffner, H.
\newblock 2009.
\newblock Plan recognition as planning.
\newblock In {\em Proceedings of the Twenty-First International Joint
  Conference on Artificial Intelligence (IJCAI-09)}.

\bibitem[\protect\citeauthoryear{Schmidt, Sridharan, and
  Goodson}{1978}]{schmidt_sridharan_goodson:aij78}
Schmidt, C.; Sridharan, N.; and Goodson, J.
\newblock 1978.
\newblock The plan recognition problem: an intersection of psychology and
  artificial intelligence.
\newblock {\em Artificial Intelligence} 11:45--83.

\bibitem[\protect\citeauthoryear{Tambe \bgroup et al.\egroup
  }{1999}]{tambe:isis-teamwork}
Tambe, M.; Adibi, J.; Al-Onaizan, Y.; Erdem, A.; Kaminka, G.~A.; Marsella,
  S.~C.; and Muslea, I.
\newblock 1999.
\newblock {B}uilding {A}gent {T}eams {U}sing an {E}xplicit {T}eamwork {M}odel
  and {L}earning.
\newblock {\em Artifical Intelligence} 110:215--239.

\bibitem[\protect\citeauthoryear{Vilain}{1990}]{vilain:aaai90}
Vilain, M.
\newblock 1990.
\newblock Getting serious about parsing plans: A grammatical analysis of plan
  recognition.
\newblock In {\em Proceedings of the Eighth National Conference on Artificial
  Intelligence},  190--197.

\bibitem[\protect\citeauthoryear{Younger}{1967}]{cyk}
Younger, D.~H.
\newblock 1967.
\newblock Recognition and parsing of context--free languages in time $n^3$.
\newblock {\em Information and Control} 10:189--208.

\end{thebibliography}

\end{document}